\pdfoutput=1

\documentclass[11pt]{article}

\usepackage[preprint]{acl}

\usepackage{times}
\usepackage{latexsym}

\usepackage[T1]{fontenc}

\usepackage[utf8]{inputenc}

\usepackage{microtype}

\usepackage{inconsolata}

\usepackage{graphicx}
\usepackage{booktabs}
\usepackage{multirow}
\usepackage{makecell}
\usepackage{amsmath}
\usepackage{amssymb}
\usepackage[most]{tcolorbox}

\usepackage{xspace}
\newcommand{\benchmark}{PEToolBench\xspace} 
\newcommand{\framework}{PEToolLLaMA\xspace}

\title{PEToolLLM: Towards Personalized Tool Learning in \\Large Language Models}

\author{Qiancheng Xu$^{1}$, Yongqi Li$^{1\dagger}$, Heming Xia$^{1}$, Fan Liu$^{2}$, Min Yang$^{3}$, Wenjie Li$^{1}$ \\
$^{1}$ The Hong Kong Polytechnic University \quad
$^{2}$ National University of Singapore \\
$^{3}$ Shenzhen Institutes of Advanced Technology, Chinese Academy of Sciences \\
\texttt{\{qiancheng.xu, he-ming.xia\}@connect.polyu.hk} \\
\texttt{liyongqi0@gmail.com} \quad
\texttt{cswjli@comp.polyu.edu.hk}
}

\begin{document}
\maketitle
\begingroup\def\thefootnote{$\dagger$}\footnotetext{Corresponding author.}\endgroup

\begin{abstract}
Tool learning has emerged as a promising direction by extending Large Language Models' (LLMs) capabilities with external tools. Existing tool learning studies primarily focus on the general-purpose tool-use capability, which addresses explicit user requirements in instructions. 
However, they overlook the importance of personalized tool-use capability, leading to an inability to handle implicit user preferences.
To address the limitation, we first formulate the task of personalized tool learning, which integrates user's interaction history towards personalized tool usage. 
To fill the gap of missing benchmarks, we construct \benchmark, featuring diverse user preferences reflected in interaction history under three distinct personalized settings, and encompassing a wide range of tool-use scenarios.
Moreover, we propose a framework \framework to adapt LLMs to the personalized tool learning task, which is trained through supervised fine-tuning and direct preference optimization.
Extensive experiments on \benchmark demonstrate the superiority of \framework over existing LLMs. 
We release our code and data at
\href{https://github.com/travis-xu/PEToolBench}{https://github.com/travis-xu/PEToolBench}.

\end{abstract}

\section{Introduction}
Large Language Models (LLMs) possess extensive knowledge and have powerful instruction-following abilities, making them effective AI assistants for tasks such as text rewriting, question answering, and code writing~\cite{zhao2023survey}.
However, they often struggle in addressing user needs in scenarios such as checking weather and booking flights. 
To address this, tool learning~\cite{10.1145/3704435,qu2024tool} has emerged as a promising solution by enabling LLMs to utilize external tools, such as real-time weather APIs and booking systems.
In this way, tool learning has extended LLMs' capabilities to tackle more complex tasks, enabling them to fulfill a wide range of user needs.

Current tool learning procedure typically begins with a user instruction, and then LLMs are required to use tools with appropriate functionalities for satisfying users' needs.
Existing tool learning methods can be categorized into in-context learning~\cite{wu-etal-2024-toolplanner,liu2025toolplanner} and fine-tuning approaches~\cite{NEURIPS2023_d842425e,wang2025toolgen}. The former approach allows LLMs to use tools by directly providing tool documentation in input but the performance is constrained by the input length.
The latter approach trains LLMs to internalize tool knowledge but struggles with tool generalization.

\begin{figure*}[!t]
    \centering
    \includegraphics[width=1.0\textwidth]{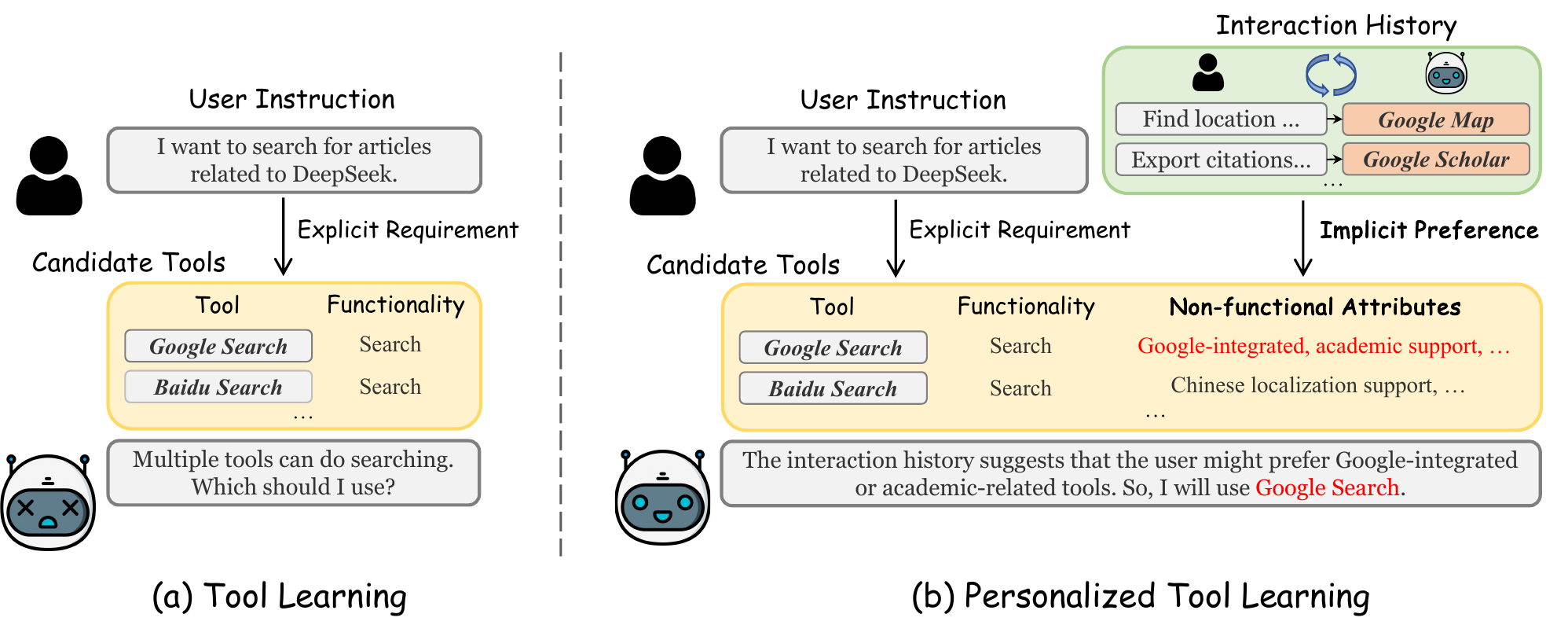}
    \caption{Comparison between (a) tool learning and (b) personalized tool learning. Personalized tool learning facilitates implicit preference comprehension and customized tool usage for individual users.}
    \label{fig:intro}
\vspace{-1em}
\end{figure*}

Despite the advancement, existing tool learning methods primarily focus on the general-purpose tool-use capability but overlook the critical role of personalization. 
In tool learning, more personalized user needs are expected to be derived from the user's previous 
tool usage 
history as a supplement to user instructions, which can help LLMs provide more customized tool-usage assistance to enhance the user experience.
As illustrated in Figure~\ref{fig:intro}, personalized tool learning is non-trivial due to the following aspects. 
1) \textbf{Implicit user preferences}. 
User preferences for tool usage are often implicitly conveyed through the user's history rather than explicitly stated in user instructions, making them difficult to understand.
For instance, when a user requests a search for articles, their preference for academic-related content needs to be inferred from previous interactions with academic tools like Google Scholar.
2) \textbf{Non-functional tool attributes}.
Since many tools have the same functionalities, user preferences cannot be effectively distinguished based solely on tool functionalities. 
This underscores the need to consider non-functional tool attributes, such as usability, integrability, and accessibility, which can better reflect user preferences. 
As shown in Figure~\ref{fig:intro}, Google Search can be distinguished from other search tools by its integration into Google’s ecosystem with Google Scholar, making it more suitable for users with academic needs.

To address the above issues, we formulate the task of personalized tool learning in LLMs, aiming at personalized tool usage for individual users.
Formally, given user instructions along with user's 
interaction history, LLMs are required to answer user instructions with tools by considering both explicit user requirements in instructions and implicit user preferences behind interaction history.


Since there is no benchmark for this task currently, we fill this gap by introducing the first personalized tool learning benchmark (\benchmark). 
Specifically, the benchmark is created through three following steps.
1) Tool Preparation.
We collect a bunch of high-quality tools from RapidAPI and then
leverage LLM to understand the functionality and non-functional attributes of each tool. 
2) Preference Construction. 
Among same-functionality tools, we construct the user's tool preferences by assigning tools with distinct non-functional attributes to different users.
3) Data Creation. 
Based on tool preference, we synthesize the user's interaction history into a sequence of tool-use interactions, each consisting of a user instruction and an LLM's tool call.
We design three personalized tool-usage settings by generating the interaction history in three types, i.e., preferred-only, rating-integrated, and chronological. 
And then we use tools not included in the interaction history to synthesize user instructions.
After rigorous filtering, we obtain 12,000 user instructions with interaction histories reflecting diverse user preferences and cover a wide range of tool-use scenarios by encompassing 7454 tools across 46 categories.


Based on the \benchmark dataset, we propose the personalized tool learning framework (\framework) to equip LLMs with personalized tool-use capability. The training process consists of two stages:
1) the supervised fine-tuning (SFT) stage, which equips LLM with foundational tool-use capability to address user needs; 2) the direct preference optimization (DPO) stage, which samples the user's preferred and non-preferred tool calls for pair-wise optimization to better align with user preferences.
We evaluate 6 distinct open-source and closed-source LLMs including the latest GPT-4o on \benchmark. 
Experimental results demonstrate that our \framework significantly outperforms the best-performing LLM across all settings with improvements even more than 50\%, showcasing its superior personalized tool-use capabilities.

In summary, our contributions are as follows.
\begin{itemize}
\item 
We are the first to formulate the task of personalized tool learning in LLMs, 
which incorporates user's interaction history to achieve personalized tool-usage assistance.

\item 
We construct the first benchmark for personalized tool learning in LLMs, \benchmark,
featuring user instructions integrated with interaction history reflecting diverse user preferences and encompassing various tools.
\item 
We propose a novel personalized tool learning framework \framework. 
Extensive experiments demonstrate that \framework significantly surpass the best-performing LLM by more than 50\%, exibiting exceptional personalized tool-use capabilities.
\end{itemize}

\section{Related Work}
\subsection{Tool Learning in LLMs}
Tool learning aims at extending the capabilities of LLMs by equipping them with external tools to solve tasks like weather inquiry, car navigation, and restaurant reservation. Existing benchmarks primarily focus on evaluating the tool learning proficiency of LLMs in addressing user instructions, from aspects such as 
tool selection and calling accuracy~\cite{xu-etal-2024-enhancing-tool,NEURIPS2024_8a75ee6d,ye-etal-2024-rotbench,wang2025mtubench}, tool planning ability~\cite{basu-etal-2024-api,wang-etal-2024-appbench,NEURIPS2024_085185ea,liu2025toolace}, and complex workflow creation~\cite{shen2025shortcutsbench,qiao2025benchmarking,fan2025workflowllm}. To improve tool-use capabilities, various strategies have been introduced, including in-context learning which enables LLMs to use tools via documentation~\cite{yuan2024easytool,shi-etal-2024-learning,qu2025from},
and fine-tuning which trains LLMs on specialized tool-use datasets~\cite{zhuang2024toolchain,chen2024advancing,chen2025learning}. However, prior studies neglect the crucial role of personalized tool usage in LLMs. This paper addresses this gap by introducing personalized tool learning, developing a comprehensive benchmark for evaluation, and proposing an optimization strategy to enhance personalized tool-use capabilities in LLMs.

\subsection{Personalization in LLMs}
The goal of personalization in LLMs is to leverage personal user data, such as historical behaviors and background information, to generate outputs that better align with the user preferences~\cite{tseng-etal-2024-two}.
Approaches such as fine-tuning~\cite{cai2025large} and prompt engineering~\cite{yuan-etal-2025-personalized}
have been explored to adapt LLMs to individual or domain-specific tasks. These approaches have been applied across various fields, including recommendation systems~\cite{lyu-etal-2024-llm}, search engines~\cite{10.1145/3589334.3645482}, education~\cite{liu2024socraticlm}, 
and dialogue generation~\cite{wang-etal-2023-target}. However, previous research has not investigated LLMs' personalization in the area of tool learning. In this work, we bridge this gap by incorporating user's interaction history to assess and enhance the LLMs' capability in providing personalized tool-usage assistance for specific users.

\begin{figure*}[!t]
    \centering
    \includegraphics[width=1.0\textwidth]{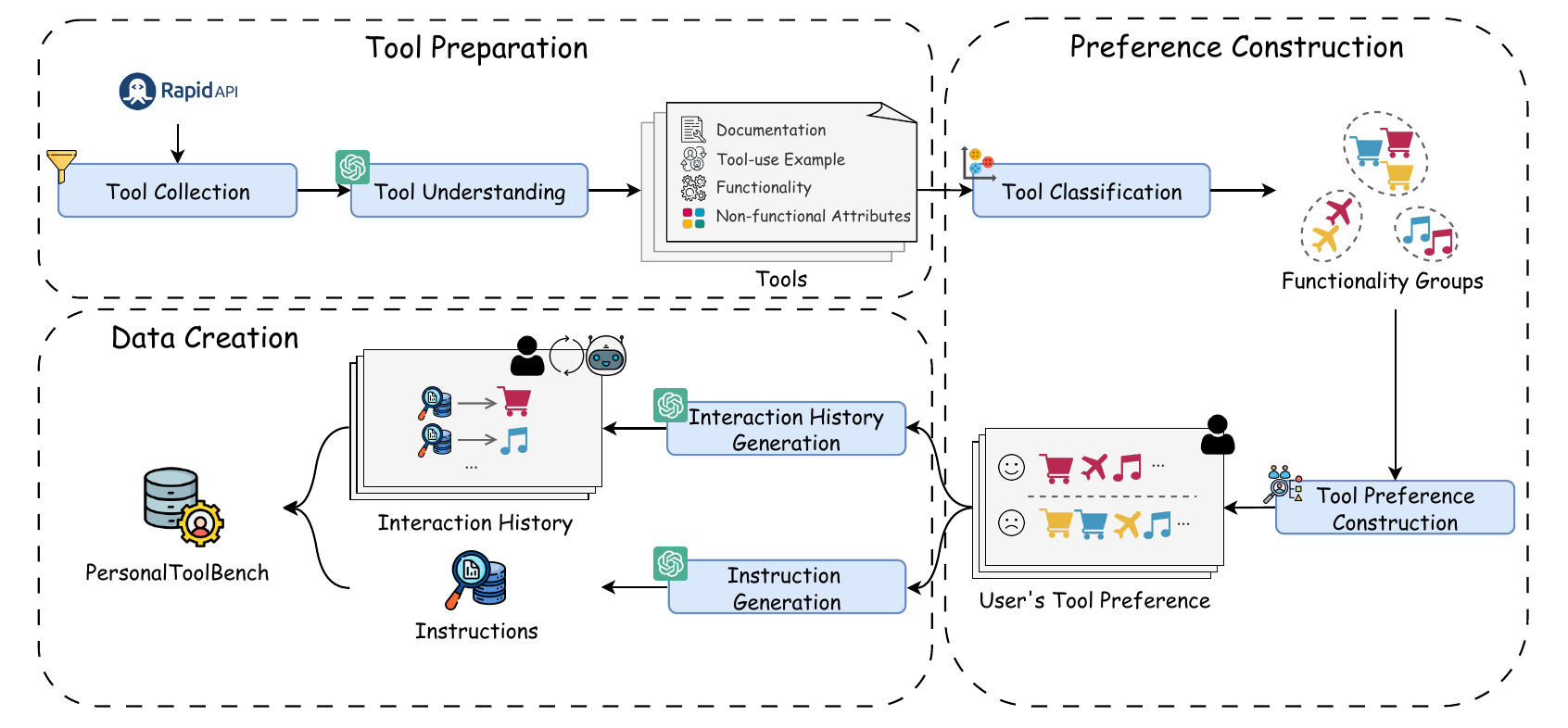}
    \caption{Illustration of the process for constructing our \benchmark.}
    \label{fig:benchmark}
\vspace{-1em}
\end{figure*}

\section{Task and Benchmark}
\subsection{Task Formulation}
\paragraph{Tool Learning}
Given an instruction $q_u$ of the user $u$, tool learning aims to generate an appropriate tool call, including the selected tool and its corresponding parameters, from a set of candidate tools. Formally, let the candidate tool set be $\mathcal{T}=\{d(t_1), d(t_2), ..., d(t_N)\}$, where $d(t_i)$ represents the documentation of tool $t_i$ and $N$ is the total number of candidate tools. The LLM is then tasked with generating a tool call $c = (t, p)$, where \(t \in \mathcal{T}\) and \(p\) denotes its parameters: 
\begin{equation}
(t,p) = LLM(q_u,\mathcal{T}).
\end{equation}

\paragraph{Personalized Tool Learning}
In personalized tool learning, we incorporate the users' interaction history alongside their instructions, enabling the LLM to generate tool calls that satisfy both the users' explicit requirements and implicit preferences.
For a user u, we define the interaction history as $\mathcal{H}_u = \{h_u^1, h_u^2, ..., h_u^M\}$, where each $h_u^i$ consists of a past user instruction $q_{u}^i$ and the corresponding tool call $c_u^i=(t_u^i,p_u^i)$, with $t_u^i$ representing the selected tool and $p_u^i$ denoting its associated parameters. Let $c_u = (t_u,p_u)$ represent the personalized tool call for user u, the personalized tool learning task can then be formulated as:
\begin{equation}
(t_u,p_u) = LLM(q_u,\mathcal{T},\mathcal{H}_u),
\end{equation}


\subsection{Benchmark Construction}
Due to the lack of real user interaction histories on tool-usage, we adopt a tool-driven approach to simulate interactions based on pre-constructed user's tool preferences. The whole process for constructing \benchmark, illustrated in Figure~\ref{fig:benchmark}, consists of three steps: tool preparation, preference construction, and data creation.
\subsubsection{Tool Preparation}

\paragraph{Tool Collection}
Following ToolBench~\cite{qin2024toolllm}, we adopt the tools from RapidAPI for our benchmark, since it offers a large-scale and diverse collection of real-world tools that can potentially address a wide range of user needs.
To ensure the quality of the collected tools, we perform strict filtering by removing: 1) outdated tools, which are marked as deprecated in RapidAPI; 2) tools with insufficient information, such as inadequate or missing tool documentation; and 3) duplicate tools, which have repeated tool names, descriptions, or category names.

\paragraph{Tool Understanding}
Since tool documentation often contains redundant and irrelevant information, directly extracting tool attributes from it can be challenging.
To address this, we 
first provide the documentation of each tool to LLM and prompt it to generate a tool-use example, including a simulated user instruction and parameters for calling the tool. 
Next, based on the tool documentation and tool-use example, the LLM is instructed to generate descriptions of the tool's functionality and non-functional attributes separately.
Besides, we include demonstrations in the prompt to help the LLM distinguish between these two attribute types.
By leveraging specific tool-use examples and demonstrations, the LLM can develop a more comprehensive understanding of each tool’s functionality and non-functional characteristics.

\subsubsection{Preference Construction}

\paragraph{Tool Classification}
To identify potential tool-usage scenarios for users, we classify tools with the same functionalities into groups.
Specifically, we first employ the Ada Embedding model~\footnote{\url{https://platform.openai.com/docs/guides/embeddings/embedding-models}.} to compute embeddings for the functionality descriptions of all tools.
Then, we apply the DBSCAN algorithm~\cite{schubert2017dbscan} to cluster these tools into multiple groups based on the similarity of their embeddings.
Within each group, the tools share the same functionality and can be applied to a specific tool-usage scenario.
To further ensure that tools within each group exhibit uniform functionality, we conduct rigorous filtering and only retain groups where tools 1) have the same input-output formats (i.e., required/optional parameters and response schema) and 2) belong to the same category (e.g., sports, music, finance).


\paragraph{Tool Preference Construction}
We leverage non-functional tool attributes to construct the user's tool preference. 
First, we randomly sample a functionality group for a user, representing a potential tool-usage scenario for interaction. Within this group, we choose a tool with specific non-functional attributes as the user's preferred tool, while the others are considered non-preferred. Using the preferred tool as a reference, we retrieve the top-$5$ tools with the most similar non-functional attributes. Similarity is computed based on the embeddings of the tools' non-functional descriptions, which are generated in the Tool Understanding phase. Through multiple iterations of sampling and retrieving, we obtain a diverse set of preferred and non-preferred tools that represent user preferences. After each iteration, we check for functionality overlap between newly retrieved tools and previously selected ones. If an overlap is detected, the tools are discarded, and the sampling process is restarted. This ensures that each tool-usage scenario is associated with only one preferred tool per user.
By following this approach, we construct diverse tool sets that align with different user preferences.

\subsubsection{Data Creation}

\paragraph{Interaction History Generation}
Based on tool preference, we leverage the LLM to construct the user's interaction history.
Specifically, for each user, we provide LLM with the user's preferred and non-preferred tools, including tool attributes and tool-use examples generated in the Tool Understanding phase.
The LLM will generate a sequence of simulated user-LLM interactions, each consisting of a user instruction and an LLM's tool call, as the user's interaction history.

We design three personalized tool-use settings by generating the interaction history in three types (illustrated in Figure~\ref{fig:history}):
1) \textit{preferred-only} history, where the tools involved in the interactions are all preferred by the user; 
2) \textit{rating-integrated} history, including both the user's preferred and non-preferred tools, with a user's binary rating for each tool-usage interaction representing the user preference, i.e., ``liked'' if the tool aligns with the user preferences, and ``disliked'' otherwise.
3) \textit{chronological} history, which organizes interactions in time order to reflect changes in user preferences over time, i.e., the more recent tool-usage interactions are more preferred by the user, while earlier interactions are less preferred.
In this way, we can present different forms of user preferences.

\paragraph{Instruction Generation}
Next, we use LLM to generate user instructions based on the user's preferred tools that are not included in the user's interaction history. 
We instruct the LLM to avoid directly generating the name of the tool in the instruction, ensuring that the user preference for the tool can only be inferred from the user's interaction history.
Each user instruction is combined with the user’s interaction history into a data instance. 

Finally, we obtain 12,000 data instances
encompassing 7,454 tools across 46 categories. We split all data into two parts: a training set comprising 9,000 instances for three personalized settings and a test set containing the rest instances.

\subsection{Benchmark Analysis}
\begin{figure}[!t]
    \centering
    \includegraphics[width=1.0\linewidth]{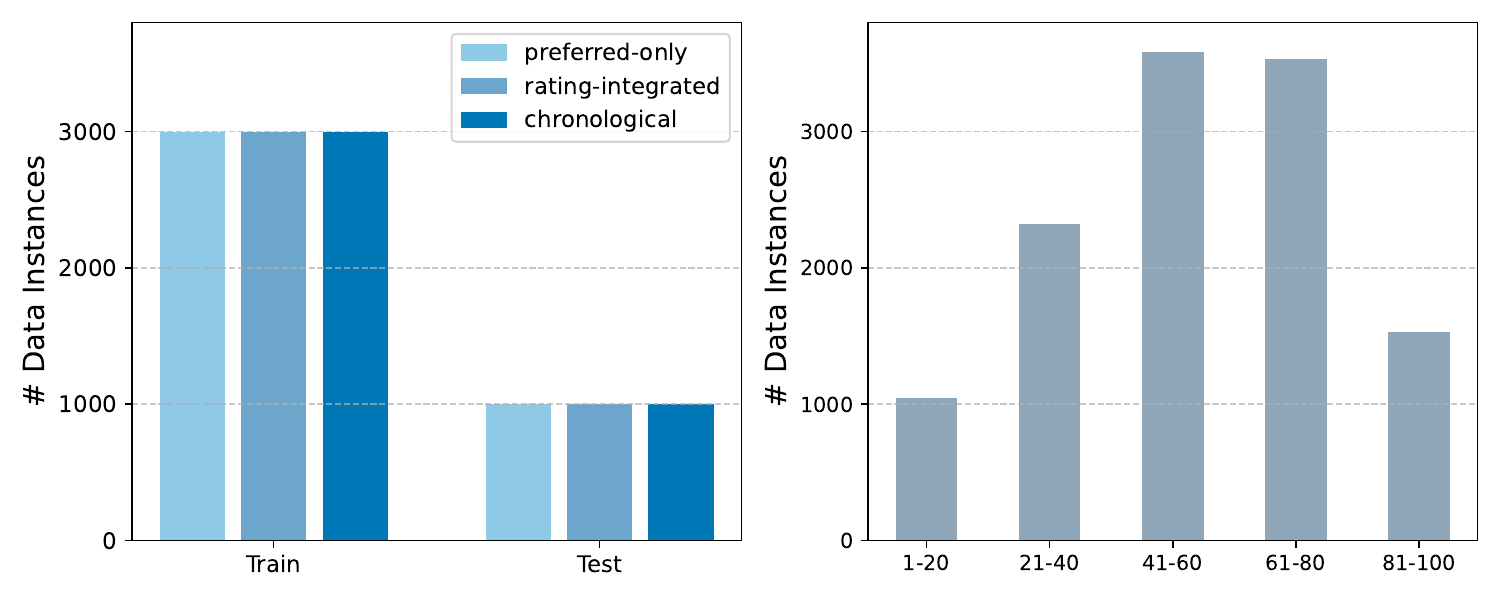}
    \caption{Statistics of data instances in three personalized settings (in the left figure) and distributions of interaction history length (in the right figure).}
    \label{fig:statistics_instances_length}
\end{figure}

\begin{figure}[tbp]
    \centering
    \includegraphics[width=1.0\linewidth]{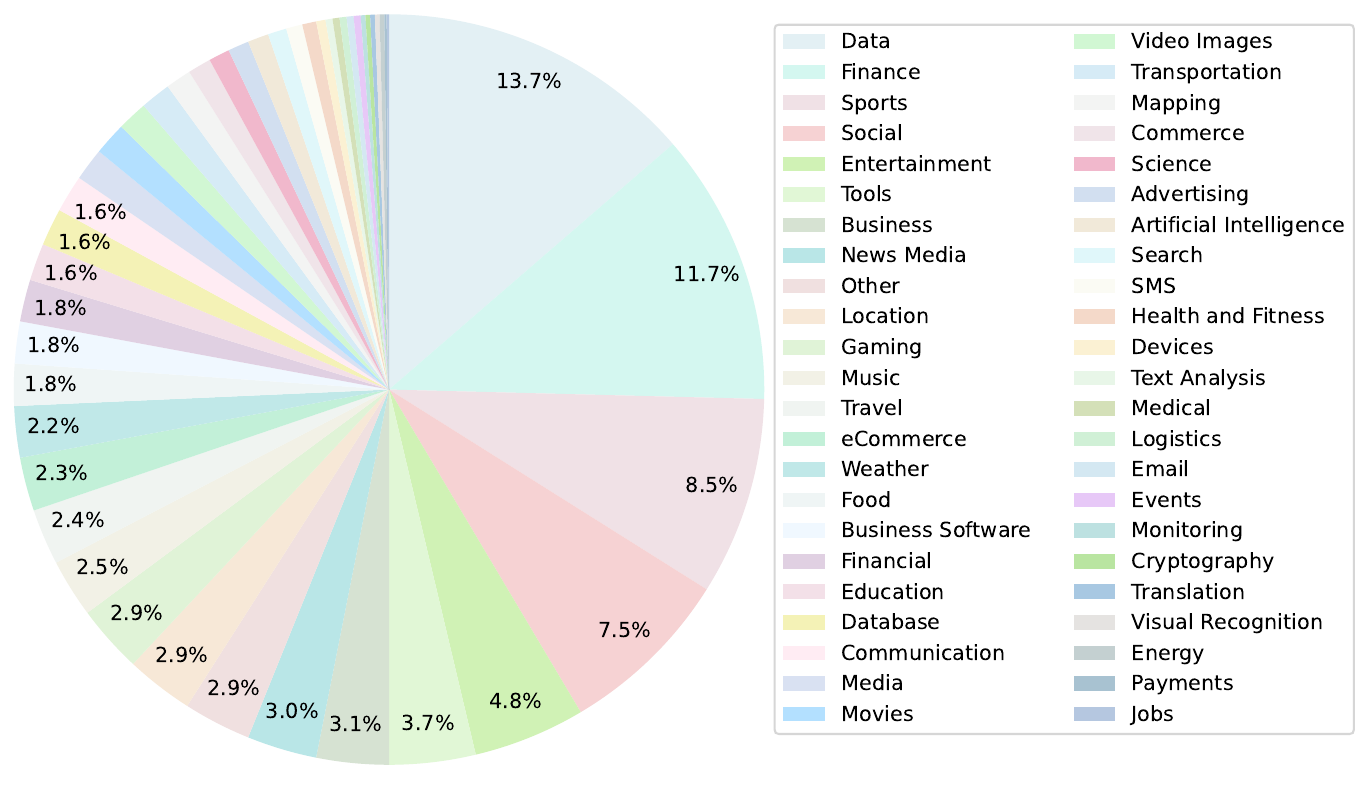}
    \caption{Distributions of tool categories.}
    \label{fig:statistics_category}
\vspace{-1em}
\end{figure}

We present the statistical information of our \benchmark in Table~\ref{fig:statistics_instances_length}, including the statistics of data instances in three settings and under varying interaction history lengths.
We also present the distribution of tool categories in Figure~\ref{fig:statistics_category}. 
Statistical information demonstrates the diversity and complexity of our dataset.

\subsection{Evaluation Metrics}
Given the user's instruction and interaction history, LLM is expected to select the appropriate tool from a candidate tool set, and then call the selected tool with corresponding parameters. Therefore, we define two metrics as follows.
\begin{itemize}
    \item Tool Accuracy (Tool Acc): The metric assesses the ability of LLM to select the appropriate tool to call. If the tool is correctly selected, the score is 1; otherwise, the score is 0.
    \item Parameter Accuracy (Param Acc): The metric assesses the ability of LLM to generate correct parameters for the tool call. If the input parameters are correctly generated, the score is 1; otherwise, the score is 0.
\end{itemize}

\begin{table*}[!t]
\centering
\caption{Evaluation results of different LLMs on \benchmark in terms of tool and parmater accuracy under settings including preferred-only, rating-integrated, chronological, and the whole data (\textit{All}). 
Bold highlights the best score among all LLMs and
\% improve represents the relative improvement achieved by our method over the previously best-performing LLM.}
\resizebox{\linewidth}{!}{
\begin{tabular}{@{}l|cc|cc|cc|cc@{}}
\toprule
\multirow{2}{*}{\textbf{Methods}} & \multicolumn{2}{c|}{\textbf{\textsc{Preferred-only}}} & \multicolumn{2}{c|}{\textbf{\textsc{Rating-integrated}}} & \multicolumn{2}{c|}{\textbf{\textsc{Chronological}}} & \multicolumn{2}{c}{\textbf{\textsc{All}}} \\ 
\cmidrule(lr){2-9}
&Tool Acc  &Param Acc  &Tool Acc &Param Acc  &Tool Acc  &Param Acc  &Tool Acc  &Param Acc  \\ \midrule
Vicuna-7B &$25.50$  &$44.80$  &$10.80$  &$57.40$  &$12.70$  &$56.00$  &$16.33$  &$52.73$  \\
Mistral-7B &$30.30$ &$55.70$  &$15.40$  &$63.20$  &$14.10$  &$64.90$  &$19.93$  &$61.27$ \\ 
Qwen2.5-7B &$40.40$  &$63.80$  &$24.80$  &$66.50$  &$24.80$  &$70.20$  &$30.00$  &$66.83$ \\ 
LLaMA3-8B &$48.10$  &$71.10$  &$26.90$  &$77.70$  &$26.60$  &$78.10$  &$33.87$  &$75.63$  \\
GPT-4o-mini &$51.80$  &$72.90$  &$38.40$  &$77.70$  &$31.20$  &$80.50$  &$40.47$  &$77.03$ \\ 
GPT-4o &$53.70$  &$77.60$  &$45.70$  &$79.60$  &$33.60$  &$81.80$  &$44.33$  &$79.67$ \\ 
\midrule
\textbf{\framework} &\bf 74.30  &\bf 87.90  &\bf 78.40  &\bf 89.70  &\bf 80.80  &\bf 91.30  &\bf 77.83  &\bf 89.63 \\ 
\% improve &38.36\% & 13.27\% &71.55\%  &12.69\%   &140.5\%  &11.61\%  &75.57\%  &12.50\%  \\ 
\textit{ w/o DPO} &71.50 &82.10 &74.20 &86.90 &77.30 &90.40 &74.33 &86.47  \\
\textit{ w/o SFT} &53.20 &61.80 &55.30 &62.40 &51.40 &61.10 &53.30 &61.77  \\
\bottomrule
\end{tabular}} 
\label{main_results}
\end{table*}

\section{Method: \framework}
To equip LLM with personalized tool-use capability, we conduct a two-stage training process: 1) personalized SFT, where LLM is fine-tuned on \benchmark to acquire fundamental proficiency in personalized tool usage, and 2) personalized DPO, where LLM is optimized on a preference dataset for better alignment with user preferences.

\paragraph{Personalized SFT.}
The first stage in our approach is Supervised Fine-Tuning (SFT), where we directly fine-tune LLM on the training set of \benchmark. Given the user's instruction $q_u$, interaction history $\mathcal{H}_u$, and the candidate tool set $\mathcal{T}$ as inputs, LLM is trained to generate the ground truth tool call \(c\). $\mathcal{H}_u$ uniformly covers all three types of user interactions to capture diverse user preferences. In this way, LLM can obtain basic personalized tool-usage experiences by understanding both the user needs and preferences.

\paragraph{Personalized DPO.}
In the second stage, we further enhance the LLM's performance through direct preference optimization (DPO)~\cite{NEURIPS2023_a85b405e}. 
Our goal is to guide the LLM to call the user's preferred tools instead of non-preferred ones.
Specifically, for each user instruction $q_u$, we collect multiple tool calls generated by LLM after the SFT stage. 
Then we select the user's preferred and non-preferred tool calls \(c_w\) and \(c_l\) based on the user's tool preference constructed in \benchmark.
\(c_w\) and \(c_l\) will be used to construct the preference dataset \(\mathcal{D}_{\text{DPO}} = \{ (x, c_w, c_l) \}\), where \(x\) denotes the input, including the user instruction $q_u$, interaction history $\mathcal{H}_u$, and the candidate tool set $\mathcal{T}$.
We then apply DPO to optimize the LLM by guiding it to generate the desired tool call \(c_w\) while avoid generating \(c_l\).
The loss function can be defined as:
\begin{equation} \label{dpo_loss}
\small
{
\mathcal{L} = -\mathbb{E} \left[ \log \sigma \left( \beta \log \frac{\pi_{\theta}(c_w \mid x)}{\pi_{\text{ref}}(c_w \mid x)} - \beta \log \frac{\pi_{\theta}(c_l \mid x)}{\pi_{\text{ref}}(c_l \mid x)} \right) \right],
}
\end{equation}
where \(\sigma\) is the logistic function and \(\beta\) is a weighting parameter that controls the deviation of the policy model $\pi_{\theta}$ (i.e., the LLM we need to optimize) from the reference model $\pi_{\text{ref}}$ (i.e., the LLM after SFT stage).
In this way, LLM can focus on generating tool calls that are more aligned with individual user preferences.

\section{Experiments}
\subsection{Setup}
\paragraph{Baselines.} 
We adopt multiple LLMs from both closed-source and open-source models to ensure a comprehensive evaluation.
For closed-source LLMs, we select two representative models: GPT-4o and GPT-4o-mini from OpenAI.
For open-source LLMs, we include a wide spectrum of models, i.e., LLaMA-3.1-8B~\cite{dubey2024llama}, QWen-2.5-7B~\citep{yang2024qwen2}, Vicuna-7B-v1.5~\cite{chiang2023vicuna} and Mistral-7B-v0.3~\cite{jiang2023mistral}.

\paragraph{Implementation details.} 
In \benchmark construction, we employ gpt-4o-mini
as the LLM for tool understanding and generation of user instructions and interaction history. 
The candidate tool set consists of three parts: the ground-truth tool along with all other tools sharing the same functionality, five tools retrieved using ToolRetriever~\cite{qin2024toolllm}, and the remaining tools that were randomly sampled.

\subsection{Main Results}
The detailed experimental results are shown in Table~\ref{main_results}. 
From the results, we can obtain the following key findings.
1) It can be observed that the performance of LLMs is generally unsatisfactory, particularly in tool accuracy with the majority failing to exceed 50\%. This indicates that current LLMs are severely limited in personalized tool-use capabilities. Additionally, the lower tool accuracy compared to parameter accuracy further suggests that personalized tool selection is more challenging than parameter configuration. This is because LLMs must account for both implicit user preferences and explicit user requirements when determining which tool to use.
2) Most LLMs perform worse in the rating-integrated and chronological settings. 
This is likely due to the inclusion of non-preferred interactions in the interaction history, which confuses LLMs and hinders their ability to accurately recognize user preferences. Notably, the chronological setting yields the lowest scores, suggesting that capturing evolving user preferences over time is even more challenging than interpreting explicit user ratings.
3) Our proposed \benchmark significantly outperforms all closed-source and open-source LLMs, demonstrating both effectiveness and robustness. It maintains strong performance, even in the two more challenging settings, by enabling the LLM to better understand diverse manifestations of user preferences and facilitate personalized tool usage.

\subsection{Ablation Study}
We conduct ablation studies to investigate the efficacy of the two-stage training process in our \framework.
First, we remove the second training stage (i.e., personalized DPO) to assess its contribution.
Then, we examine the impact of the SFT stage by directly conducting DPO training on the initial LLaMA3-8B model.
Table~\ref{main_results} reports the performance on the test set of \benchmark in all three settings.
The results indicate that the SFT stage is crucial for personalized tool learning performance, as it endows the model with fundamental tool usage and personalization capabilities. Removing the DPO stage results in a slight performance drop, suggesting that it can further refine the tool usage alignment with user preferences.

\subsection{In-depth Analysis}

\begin{figure}[tbp]
    \centering
    \includegraphics[width=1.0\linewidth]{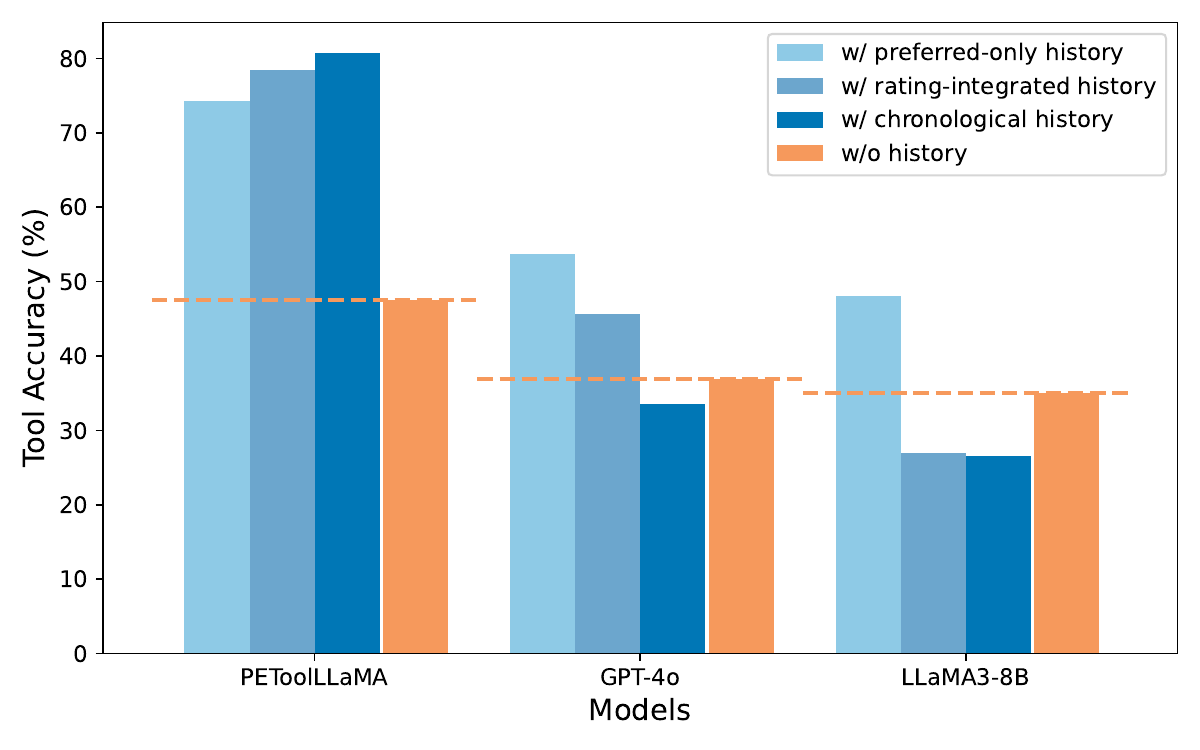}
    \caption{Performance comparison of tool accuracy when provided with and without interaction history.}
    \label{fig:scores_wo_history}
\vspace{-1em}
\end{figure}

\begin{figure}[!t]
    \centering
    \includegraphics[width=1.0\linewidth]{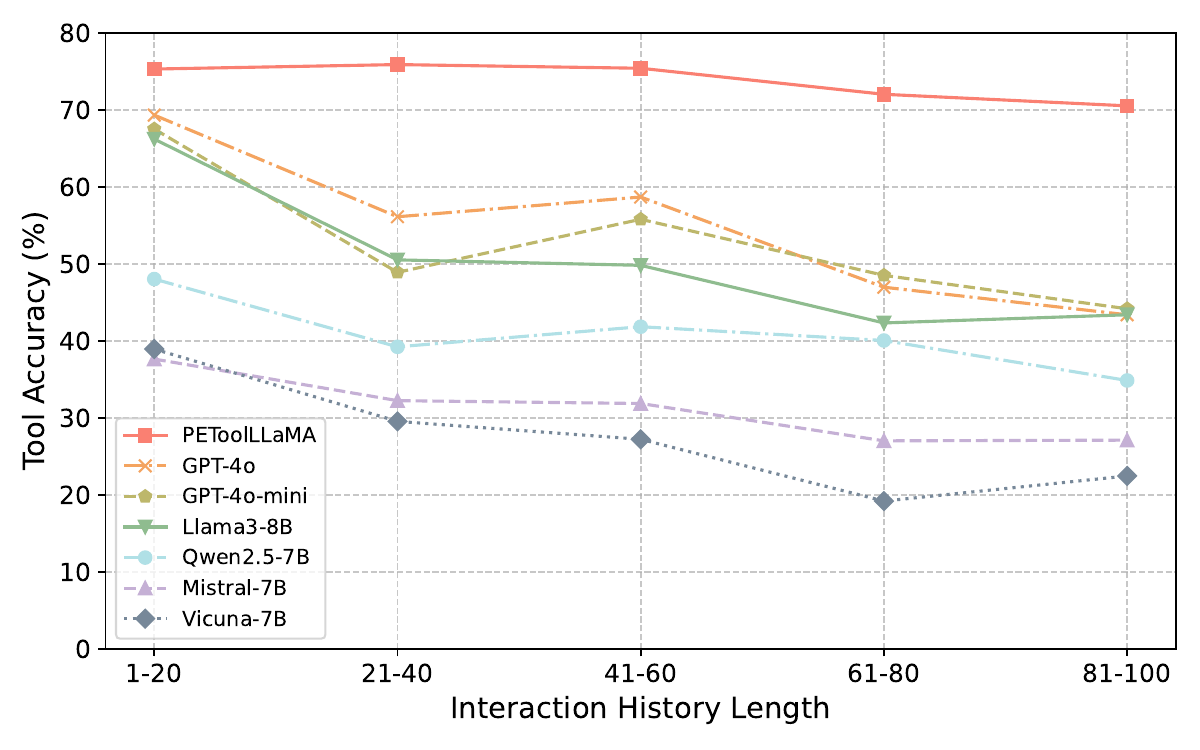}
    \caption{Performance comparison of tool accuracy on different interaction history length in the preferred-only setting.}
    \label{fig:scores_length}
\vspace{-1em}
\end{figure}

\begin{table*}[t]
\centering
\caption{The percentage (\%) of different error types in LLMs on the test set of \benchmark under preferred-only and chronological settings. IF, TH, TFM, TPM, PNM, PVM stand for Invalid Format, Tool Hallucination, Tool Functionality Mismatch, Tool Preference Mismatch, Parameter Name Mismatch and Parameter Value Mismatch errors, respectively. 
} 
\resizebox{0.87\linewidth}{!}
{
\begin{tabular}{@{}l|c|ccc|cc|c|ccc|cc@{}}
\toprule
\multirow{2}{*}{\textbf{Models}} & \multicolumn{6}{c|}{\textbf{\textsc{Preferred-only}}} & \multicolumn{6}{c}{\textbf{\textsc{Chronological}}} \\ 
\cmidrule(lr){2-13}
&IF  &TH  &TFM &TPM  &PNM  &PVM &IF  &TH  &TFM &TPM  &PNM  &PVM \\ 
 \midrule
Qwen2.5-7B &$10.9$  &$3.6$  &$19.6$  &$25.5$  &$10.4$  &$14.9$ &$11.2$  &$2.3$  &$7.1$  &$54.6$  &$5.4$  &$13.2$  \\ 
LLaMA3-8B &$2.5$  &$5.3$  &$19.3$  &$24.8$  &$11.4$  &$15.0$ &$2.5$  &$3.7$  &$2.8$  &$64.4$  &$6.8$  &$12.6$ \\
GPT-4o-mini &\bf 0.1  &$3.5$  &$20.1$  &$24.4$  &$10.6$  &$16.4$ &\bf 0.0  &$1.5$  &$6.9$  &$60.2$  &$6.2$  &$13.3$ \\ 
GPT-4o &$0.5$  &\bf 1.5 &$20.0$  &$30.3$  &$7.9$  &$14.0$ &$1.3$  &\bf 1.1 &$6.6$  &$57.3$  &\bf 4.7  &$12.2$\\ 
\textbf{\framework} &$0.6$  &$3.4$  &\bf 9.3  &\bf 12.0  &\bf 7.6  &\bf 3.9 &$0.5$  &$1.8$  &\bf 6.5  &\bf 10.4  &$5.1$  &\bf 3.1 \\ 
\bottomrule
\end{tabular}} 
\label{error_results}
\end{table*}


\paragraph{Analysis on the impact of interaction history.}
To investigate the impact of interaction history on LLM performance, we remove the interaction history from the inputs and provide only the user instructions with candidate tools set to conduct our experiments. The results are presented in Figure~\ref{fig:scores_wo_history}.
From the results, we can observe that both closed-source and open-source LLMs experience varying degrees of performance degradation without interaction history, compared to when provided with preferred-only history. This suggests that interaction history only containing the user's preferred tools can help the LLM effectively infer user preferences. On the other hand, we find that LLMs perform better in the absence of interaction history than with chronological history. This indicates that including both preferred and non-preferred tools can interfere with the LLM's understanding of user preferences, thus hindering its personalization capabilities.
In contrast, our \framework consistently improves performance across all three types of interaction history compared to the no-history setting. This demonstrates that our method enables LLM to effectively recognize different forms of user preferences from the interaction history.

\paragraph{Analysis on interaction history length.}
To evaluate the performance of LLMs under varying interaction history lengths, we break down the tool accuracy scores of LLMs based on the number of interactions in the history under the preferred-only setting.
As shown in Figure~\ref{fig:scores_length}, the performance of both closed-source and open-source LLMs deteriorates as interaction history length increases.
This is because a longer interaction history makes it more challenging for the LLM to identify the historical preferences relevant to identify relevant historical preferences in relation to the user’s current context.
In contrast, our \framework significantly outperforms all LLMs and maintains strong, consistent performance even as interaction history grows. This demonstrates that our method enables LLMs to effectively extract and utilize user preferences from complex historical data.

\subsection{Error Analysis}
We further conduct an error analysis to investigate the issues leading to incorrect tool calls in two personalized settings. 
We categorize the errors into six types:
1) Invalid Format. The tool call generated by the LLMs does not follow the expected JSON format. 
2) Tool Hallucination. The LLM generates a tool that does not exist in the given candidate tool set, which is a common hallucination issue in LLMs.
3) Tool Functionality Mismatch. The selected tool lacks the necessary functionality to fulfill the user’s requirements.
4) Tool Preference Mismatch. The selected tool has the correct functionality but is not preferred by the user.
5) Parameter Name Mismatch. The tool call contains missing or incorrect parameter names.
6) Parameter Value Mismatch. The parameter names are correctly generated, but the parameter values do not match the ground truth.

From the results in Table~\ref{error_results}, we observe that most LLMs perform worst in Tool Preference Mismatch, particularly in the chronological setting, where the error rate exceeds 50\%. This suggests that identifying user preferences from the interaction history is highly challenging, especially when preferences change over time, leading to significant model misinterpretation. In contrast, our \framework significantly reduces the error rate in Tool Preference Mismatch, demonstrating its effectiveness in capturing implicit user preferences. Additionally, the reduction in Tool Functionality Mismatch and Parameter Value Mismatch errors suggests that our method enhances LLMs' fundamental tool-usage ability, improving their handling of explicit user requirements.
Furthermore, \framework achieves low error rates in Invalid Format and Tool Hallucination, comparable to closed-source LLMs, highlighting its strong instruction-following capabilities.

\section{Conclusion and Future Work}
In this paper, we advanced general-purpose tool-use LLMs into personalized tool-use LLMs, aiming to provide users with customized tool-usage assistance. We formulate the task of personalized tool learning and identify the goal of leveraging user's interaction history to achieve implicit preference understanding and personalized tool calling. 
For training and evaluation, we construct the first \benchmark benchmark, featuring diverse users’ interaction history in three types. 
We also propose a novel personalized framework \framework conducted under a two-stage training process to endow LLMs with personalized tool-use capabilities.
Extensive experiments on \benchmark demonstrate that \framework consistently surpasses existing baselines, effectively meeting user requirements and preferences.
We believe that the task, benchmark, and framework for personalized tool learning will broaden the research scope, introduce new challenges and inspire novel methods. 

In the future, we aim to enhance this work from the following dimensions.
1) We plan to explore more heterogeneous personal user data beyond interaction history, such as user profiles or personas. This will allow us to reflect user preferences from multiple dimensions, providing a more comprehensive evaluation on the personalized tool-use capabilities of LLMs.
2) Currently, our work is limited to tool-usage scenarios involving a single tool. In the future, we intend to expand to more complex personalized tool-usage, such as multi-tool scenarios. These scenarios will require LLMs to perform personalized tool planning and engage in multi-round tool calling to address user needs effectively.

\section*{Limitations}
1) Due to the lack of real user interaction histories on tool usage, we utilize LLM to synthesize such data. However, this approach may compromise the authenticity and reliability of the data, which is a common challenge in data synthesis methods.
To mitigate this issue, we incorporate pre-constructed user preference information into the data generation process. This strategy helps guide LLM in generating contextually relevant outputs, thereby improving the quality and consistency of the synthesized data.
2) In real-world scenarios, tools have multiple dimensions of attributes. However, due to the limited information contained in tool documentation, it is difficult to identify and fully exploit all possible tool attributes. Fortunately, the attributes we have obtained are sufficient to differentiate between tools, enabling us to effectively construct user preferences.

\section*{Ethics Statement}
The dataset used in our work is derived from publicly available sources and generated through interactions with LLMs in English. Since the user interaction histories in our study are entirely simulated, user privacy is fully protected, and no real personal information is included in the dataset. Furthermore, all scientific artifacts used in this research are publicly accessible for academic purposes under permissive licenses, and their use in this paper complies with their intended purposes. Given these considerations, we believe our research adheres to the ethical standards of the conference.

\bibliography{acl_latex}

\begin{thebibliography}{38}
\providecommand{\natexlab}[1]{#1}

\bibitem[{Basu et~al.(2024)Basu, Abdelaziz, Chaudhury, Dan, Crouse, Munawar, Austel, Kumaravel, Muthusamy, Kapanipathi, and Lastras}]{basu-etal-2024-api}
Kinjal Basu, Ibrahim Abdelaziz, Subhajit Chaudhury, Soham Dan, Maxwell Crouse, Asim Munawar, Vernon Austel, Sadhana Kumaravel, Vinod Muthusamy, Pavan Kapanipathi, and Luis Lastras. 2024.
\newblock {API}-{BLEND}: A comprehensive corpora for training and benchmarking {API} {LLM}s.
\newblock In \emph{Proceedings of the 62nd Annual Meeting of the Association for Computational Linguistics (Volume 1: Long Papers)}, pages 12859--12870. Association for Computational Linguistics.

\bibitem[{Cai et~al.(2025)Cai, Li, Wang, ZHU, Shen, Li, and Chua}]{cai2025large}
Hongru Cai, Yongqi Li, Wenjie Wang, Fengbin ZHU, Xiaoyu Shen, Wenjie Li, and Tat-Seng Chua. 2025.
\newblock Large language models empowered personalized web agents.
\newblock In \emph{THE WEB CONFERENCE 2025}.

\bibitem[{Chen et~al.(2025)Chen, Zhang, Cong, Guo, Wu, Lin, Feng, and Wang}]{chen2025learning}
Guoxin Chen, Zhong Zhang, Xin Cong, Fangda Guo, Yesai Wu, Yankai Lin, Wenzheng Feng, and Yasheng Wang. 2025.
\newblock Learning evolving tools for large language models.
\newblock In \emph{The Thirteenth International Conference on Learning Representations}.

\bibitem[{Chen et~al.(2024)Chen, Wang, Wu, Chen, Xu, Luo, Zhang, and Zhang}]{chen2024advancing}
Sijia Chen, Yibo Wang, Yi-Feng Wu, Qing-Guo Chen, Zhao Xu, Weihua Luo, Kaifu Zhang, and Lijun Zhang. 2024.
\newblock Advancing tool-augmented large language models: Integrating insights from errors in inference trees.
\newblock In \emph{The Thirty-eighth Annual Conference on Neural Information Processing Systems}.

\bibitem[{Chiang et~al.(2023)Chiang, Li, Lin, Sheng, Wu, Zhang, Zheng, Zhuang, Zhuang, Gonzalez et~al.}]{chiang2023vicuna}
Wei-Lin Chiang, Zhuohan Li, Zi~Lin, Ying Sheng, Zhanghao Wu, Hao Zhang, Lianmin Zheng, Siyuan Zhuang, Yonghao Zhuang, Joseph~E Gonzalez, et~al. 2023.
\newblock Vicuna: An open-source chatbot impressing gpt-4 with 90\%* chatgpt quality.
\newblock \emph{See https://vicuna. lmsys. org (accessed 14 April 2023)}, 2(3):6.

\bibitem[{Dubey et~al.(2024)Dubey, Jauhri, Pandey, Kadian, Al-Dahle, Letman, Mathur, Schelten, Yang, Fan et~al.}]{dubey2024llama}
Abhimanyu Dubey, Abhinav Jauhri, Abhinav Pandey, Abhishek Kadian, Ahmad Al-Dahle, Aiesha Letman, Akhil Mathur, Alan Schelten, Amy Yang, Angela Fan, et~al. 2024.
\newblock The llama 3 herd of models.
\newblock \emph{arXiv preprint arXiv:2407.21783}.

\bibitem[{Fan et~al.(2025)Fan, Cong, Fu, Zhang, Zhang, Liu, Wu, Lin, Liu, and Sun}]{fan2025workflowllm}
Shengda Fan, Xin Cong, Yuepeng Fu, Zhong Zhang, Shuyan Zhang, Yuanwei Liu, Yesai Wu, Yankai Lin, Zhiyuan Liu, and Maosong Sun. 2025.
\newblock Workflow{LLM}: Enhancing workflow orchestration capability of large language models.
\newblock In \emph{The Thirteenth International Conference on Learning Representations}.

\bibitem[{Jiang et~al.(2023)Jiang, Sablayrolles, Mensch, Bamford, Chaplot, Casas, Bressand, Lengyel, Lample, Saulnier et~al.}]{jiang2023mistral}
Albert~Q Jiang, Alexandre Sablayrolles, Arthur Mensch, Chris Bamford, Devendra~Singh Chaplot, Diego de~las Casas, Florian Bressand, Gianna Lengyel, Guillaume Lample, Lucile Saulnier, et~al. 2023.
\newblock Mistral 7b.
\newblock \emph{arXiv preprint arXiv:2310.06825}.

\bibitem[{Liu et~al.(2024)Liu, Huang, Xiao, Sha, Wu, Liu, Wang, and Chen}]{liu2024socraticlm}
Jiayu Liu, Zhenya Huang, Tong Xiao, Jing Sha, Jinze Wu, Qi~Liu, Shijin Wang, and Enhong Chen. 2024.
\newblock Socratic{LM}: Exploring socratic personalized teaching with large language models.
\newblock In \emph{The Thirty-eighth Annual Conference on Neural Information Processing Systems}.

\bibitem[{Liu et~al.(2025{\natexlab{a}})Liu, Zeng, Huang, xinlong hao, Yu, Li, Wang, Gan, Liu, Yu, WANG, Wang, Ning, Hou, Wang, Wu, Xinzhi, Liu, Wang, Tang, Tu, Shang, Jiang, Tang, Lian, Liu, and Chen}]{liu2025toolace}
Weiwen Liu, Xingshan Zeng, Xu~Huang, xinlong hao, Shuai Yu, Dexun Li, Shuai Wang, Weinan Gan, Zhengying Liu, Yuanqing Yu, Zezhong WANG, Yuxian Wang, Wu~Ning, Yutai Hou, Bin Wang, Chuhan Wu, Wang Xinzhi, Yong Liu, Yasheng Wang, Duyu Tang, Dandan Tu, Lifeng Shang, Xin Jiang, Ruiming Tang, Defu Lian, Qun Liu, and Enhong Chen. 2025{\natexlab{a}}.
\newblock Tool{ACE}: Enhancing function calling with accuracy, complexity, and diversity.
\newblock In \emph{The Thirteenth International Conference on Learning Representations}.

\bibitem[{Liu et~al.(2025{\natexlab{b}})Liu, Peng, Cao, Bo, Zhang, Zhang, Cheng, Wang, Yin, and Du}]{liu2025toolplanner}
Yanming Liu, Xinyue Peng, Jiannan Cao, Shi Bo, Yuwei Zhang, Xuhong Zhang, Sheng Cheng, Xun Wang, Jianwei Yin, and Tianyu Du. 2025{\natexlab{b}}.
\newblock Tool-planner: Task planning with clusters across multiple tools.
\newblock In \emph{The Thirteenth International Conference on Learning Representations}.

\bibitem[{Lyu et~al.(2024)Lyu, Jiang, Zeng, Xia, Wang, Zhang, Chen, Leung, Tang, and Luo}]{lyu-etal-2024-llm}
Hanjia Lyu, Song Jiang, Hanqing Zeng, Yinglong Xia, Qifan Wang, Si~Zhang, Ren Chen, Chris Leung, Jiajie Tang, and Jiebo Luo. 2024.
\newblock {LLM}-rec: Personalized recommendation via prompting large language models.
\newblock In \emph{Findings of the Association for Computational Linguistics: NAACL 2024}, pages 583--612. Association for Computational Linguistics.

\bibitem[{Qiao et~al.(2025)Qiao, Fang, Qiu, Wang, Zhang, Jiang, Xie, Huang, and Chen}]{qiao2025benchmarking}
Shuofei Qiao, Runnan Fang, Zhisong Qiu, Xiaobin Wang, Ningyu Zhang, Yong Jiang, Pengjun Xie, Fei Huang, and Huajun Chen. 2025.
\newblock Benchmarking agentic workflow generation.
\newblock In \emph{The Thirteenth International Conference on Learning Representations}.

\bibitem[{Qin et~al.(2024{\natexlab{a}})Qin, Hu, Lin, Chen, Ding, Cui, Zeng, Zhou, Huang, Xiao, Han, Fung, Su, Wang, Qian, Tian, Zhu, Liang, Shen, Xu, Zhang, Ye, Li, Tang, Yi, Zhu, Dai, Yan, Cong, Lu, Zhao, Huang, Yan, Han, Sun, Li, Phang, Yang, Wu, Ji, Li, Liu, and Sun}]{10.1145/3704435}
Yujia Qin, Shengding Hu, Yankai Lin, Weize Chen, Ning Ding, Ganqu Cui, Zheni Zeng, Xuanhe Zhou, Yufei Huang, Chaojun Xiao, Chi Han, Yi~Ren Fung, Yusheng Su, Huadong Wang, Cheng Qian, Runchu Tian, Kunlun Zhu, Shihao Liang, Xingyu Shen, Bokai Xu, Zhen Zhang, Yining Ye, Bowen Li, Ziwei Tang, Jing Yi, Yuzhang Zhu, Zhenning Dai, Lan Yan, Xin Cong, Yaxi Lu, Weilin Zhao, Yuxiang Huang, Junxi Yan, Xu~Han, Xian Sun, Dahai Li, Jason Phang, Cheng Yang, Tongshuang Wu, Heng Ji, Guoliang Li, Zhiyuan Liu, and Maosong Sun. 2024{\natexlab{a}}.
\newblock Tool learning with foundation models.
\newblock \emph{ACM Comput. Surv.}, 57(4).

\bibitem[{Qin et~al.(2024{\natexlab{b}})Qin, Liang, Ye, Zhu, Yan, Lu, Lin, Cong, Tang, Qian, Zhao, Hong, Tian, Xie, Zhou, Gerstein, dahai li, Liu, and Sun}]{qin2024toolllm}
Yujia Qin, Shihao Liang, Yining Ye, Kunlun Zhu, Lan Yan, Yaxi Lu, Yankai Lin, Xin Cong, Xiangru Tang, Bill Qian, Sihan Zhao, Lauren Hong, Runchu Tian, Ruobing Xie, Jie Zhou, Mark Gerstein, dahai li, Zhiyuan Liu, and Maosong Sun. 2024{\natexlab{b}}.
\newblock Tool{LLM}: Facilitating large language models to master 16000+ real-world {API}s.
\newblock In \emph{The Twelfth International Conference on Learning Representations}.

\bibitem[{Qu et~al.(2024)Qu, Dai, Wei, Cai, Wang, Yin, Xu, and Wen}]{qu2024tool}
Changle Qu, Sunhao Dai, Xiaochi Wei, Hengyi Cai, Shuaiqiang Wang, Dawei Yin, Jun Xu, and Ji-Rong Wen. 2024.
\newblock Tool learning with large language models: A survey.
\newblock \emph{arXiv preprint arXiv:2405.17935}.

\bibitem[{Qu et~al.(2025)Qu, Dai, Wei, Cai, Wang, Yin, Xu, and Wen}]{qu2025from}
Changle Qu, Sunhao Dai, Xiaochi Wei, Hengyi Cai, Shuaiqiang Wang, Dawei Yin, Jun Xu, and Ji-Rong Wen. 2025.
\newblock From exploration to mastery: Enabling {LLM}s to master tools via self-driven interactions.
\newblock In \emph{The Thirteenth International Conference on Learning Representations}.

\bibitem[{Rafailov et~al.(2023)Rafailov, Sharma, Mitchell, Manning, Ermon, and Finn}]{NEURIPS2023_a85b405e}
Rafael Rafailov, Archit Sharma, Eric Mitchell, Christopher~D Manning, Stefano Ermon, and Chelsea Finn. 2023.
\newblock Direct preference optimization: Your language model is secretly a reward model.
\newblock In \emph{Advances in Neural Information Processing Systems}, pages 53728--53741. Curran Associates, Inc.

\bibitem[{Schick et~al.(2023)Schick, Dwivedi-Yu, Dessi, Raileanu, Lomeli, Hambro, Zettlemoyer, Cancedda, and Scialom}]{NEURIPS2023_d842425e}
Timo Schick, Jane Dwivedi-Yu, Roberto Dessi, Roberta Raileanu, Maria Lomeli, Eric Hambro, Luke Zettlemoyer, Nicola Cancedda, and Thomas Scialom. 2023.
\newblock Toolformer: Language models can teach themselves to use tools.
\newblock In \emph{Advances in Neural Information Processing Systems}, volume~36, pages 68539--68551. Curran Associates, Inc.

\bibitem[{Schubert et~al.(2017)Schubert, Sander, Ester, Kriegel, and Xu}]{schubert2017dbscan}
Erich Schubert, J{\"o}rg Sander, Martin Ester, Hans~Peter Kriegel, and Xiaowei Xu. 2017.
\newblock Dbscan revisited, revisited: why and how you should (still) use dbscan.
\newblock \emph{ACM Transactions on Database Systems (TODS)}, 42(3):1--21.

\bibitem[{SHEN et~al.(2025)SHEN, Li, Meng, Cai, Qi, Zhang, Xu, and Ma}]{shen2025shortcutsbench}
Haiyang SHEN, Yue Li, Desong Meng, Dongqi Cai, Sheng Qi, Li~Zhang, Mengwei Xu, and Yun Ma. 2025.
\newblock Shortcutsbench: A large-scale real-world benchmark for {API}-based agents.
\newblock In \emph{The Thirteenth International Conference on Learning Representations}.

\bibitem[{Shen et~al.(2024)Shen, Song, Tan, Zhang, Ren, Yuan, Lu, Li, and Zhuang}]{NEURIPS2024_085185ea}
Yongliang Shen, Kaitao Song, Xu~Tan, Wenqi Zhang, Kan Ren, Siyu Yuan, Weiming Lu, Dongsheng Li, and Yueting Zhuang. 2024.
\newblock Taskbench: Benchmarking large language models for task automation.
\newblock In \emph{Advances in Neural Information Processing Systems}, volume~37, pages 4540--4574. Curran Associates, Inc.

\bibitem[{Shi et~al.(2024)Shi, Gao, Chen, Feng, Yan, Shi, Yin, Ren, Verberne, and Ren}]{shi-etal-2024-learning}
Zhengliang Shi, Shen Gao, Xiuyi Chen, Yue Feng, Lingyong Yan, Haibo Shi, Dawei Yin, Pengjie Ren, Suzan Verberne, and Zhaochun Ren. 2024.
\newblock Learning to use tools via cooperative and interactive agents.
\newblock In \emph{Findings of the Association for Computational Linguistics: EMNLP 2024}, pages 10642--10657. Association for Computational Linguistics.

\bibitem[{Tseng et~al.(2024)Tseng, Huang, Hsiao, Chen, Huang, Meng, and Chen}]{tseng-etal-2024-two}
Yu-Min Tseng, Yu-Chao Huang, Teng-Yun Hsiao, Wei-Lin Chen, Chao-Wei Huang, Yu~Meng, and Yun-Nung Chen. 2024.
\newblock Two tales of persona in {LLM}s: A survey of role-playing and personalization.
\newblock In \emph{Findings of the Association for Computational Linguistics: EMNLP 2024}, pages 16612--16631. Association for Computational Linguistics.

\bibitem[{Wang et~al.(2024{\natexlab{a}})Wang, Wang, Xue, Xia, Cao, Liu, Pan, and Wong}]{wang-etal-2024-appbench}
Hongru Wang, Rui Wang, Boyang Xue, Heming Xia, Jingtao Cao, Zeming Liu, Jeff~Z. Pan, and Kam-Fai Wong. 2024{\natexlab{a}}.
\newblock {A}pp{B}ench: Planning of multiple {API}s from various {APP}s for complex user instruction.
\newblock In \emph{Proceedings of the 2024 Conference on Empirical Methods in Natural Language Processing}, pages 15322--15336. Association for Computational Linguistics.

\bibitem[{Wang et~al.(2023)Wang, Cheng, Lin, Leong, and Li}]{wang-etal-2023-target}
Jian Wang, Yi~Cheng, Dongding Lin, Chak Leong, and Wenjie Li. 2023.
\newblock Target-oriented proactive dialogue systems with personalization: Problem formulation and dataset curation.
\newblock In \emph{Proceedings of the 2023 Conference on Empirical Methods in Natural Language Processing}, pages 1132--1143. Association for Computational Linguistics.

\bibitem[{Wang et~al.(2024{\natexlab{b}})Wang, Zerun, Li, Zhang, Chen, Chen, and Le}]{NEURIPS2024_8a75ee6d}
Jize Wang, Ma~Zerun, Yining Li, Songyang Zhang, Cailian Chen, Kai Chen, and Xinyi Le. 2024{\natexlab{b}}.
\newblock Gta: A benchmark for general tool agents.
\newblock In \emph{Advances in Neural Information Processing Systems}, volume~37, pages 75749--75790. Curran Associates, Inc.

\bibitem[{Wang et~al.(2025{\natexlab{a}})Wang, Wu, Wang, Liu, Song, Peng, Deng, Zhang, JiakaiWang, Peng, Zhang, Guo, Zhang, Su, and Zheng}]{wang2025mtubench}
Pei Wang, Yanan Wu, Noah Wang, Jiaheng Liu, Xiaoshuai Song, Z.Y. Peng, Ken Deng, Chenchen Zhang, JiakaiWang, Junran Peng, Ge~Zhang, Hangyu Guo, Zhaoxiang Zhang, Wenbo Su, and Bo~Zheng. 2025{\natexlab{a}}.
\newblock {MTU}-bench: A multi-granularity tool-use benchmark for large language models.
\newblock In \emph{The Thirteenth International Conference on Learning Representations}.

\bibitem[{Wang et~al.(2025{\natexlab{b}})Wang, Han, Ji, Wang, Baldwin, and Li}]{wang2025toolgen}
Renxi Wang, Xudong Han, Lei Ji, Shu Wang, Timothy Baldwin, and Haonan Li. 2025{\natexlab{b}}.
\newblock Toolgen: Unified tool retrieval and calling via generation.
\newblock In \emph{The Thirteenth International Conference on Learning Representations}.

\bibitem[{Wu et~al.(2024)Wu, Liu, Luan, and Wang}]{wu-etal-2024-toolplanner}
Qinzhuo Wu, Wei Liu, Jian Luan, and Bin Wang. 2024.
\newblock {T}ool{P}lanner: A tool augmented {LLM} for multi granularity instructions with path planning and feedback.
\newblock In \emph{Proceedings of the 2024 Conference on Empirical Methods in Natural Language Processing}, pages 18315--18339. Association for Computational Linguistics.

\bibitem[{Xu et~al.(2024)Xu, Li, Xia, and Li}]{xu-etal-2024-enhancing-tool}
Qiancheng Xu, Yongqi Li, Heming Xia, and Wenjie Li. 2024.
\newblock Enhancing tool retrieval with iterative feedback from large language models.
\newblock In \emph{Findings of the Association for Computational Linguistics: EMNLP 2024}, pages 9609--9619. Association for Computational Linguistics.

\bibitem[{Yang et~al.(2024)Yang, Yang, Zhang, Hui, Zheng, Yu, Li, Liu, Huang, Wei et~al.}]{yang2024qwen2}
An~Yang, Baosong Yang, Beichen Zhang, Binyuan Hui, Bo~Zheng, Bowen Yu, Chengyuan Li, Dayiheng Liu, Fei Huang, Haoran Wei, et~al. 2024.
\newblock Qwen2. 5 technical report.
\newblock \emph{arXiv preprint arXiv:2412.15115}.

\bibitem[{Ye et~al.(2024)Ye, Wu, Gao, Huang, Li, Li, Fan, Zhang, Gui, and Huang}]{ye-etal-2024-rotbench}
Junjie Ye, Yilong Wu, Songyang Gao, Caishuang Huang, Sixian Li, Guanyu Li, Xiaoran Fan, Qi~Zhang, Tao Gui, and Xuanjing Huang. 2024.
\newblock {R}o{TB}ench: A multi-level benchmark for evaluating the robustness of large language models in tool learning.
\newblock In \emph{Proceedings of the 2024 Conference on Empirical Methods in Natural Language Processing}, pages 313--333. Association for Computational Linguistics.

\bibitem[{Yuan et~al.(2025)Yuan, Sun, Li, Wang, Cao, and Li}]{yuan-etal-2025-personalized}
Ruifeng Yuan, Shichao Sun, Yongqi Li, Zili Wang, Ziqiang Cao, and Wenjie Li. 2025.
\newblock Personalized large language model assistant with evolving conditional memory.
\newblock In \emph{Proceedings of the 31st International Conference on Computational Linguistics}, pages 3764--3777. Association for Computational Linguistics.

\bibitem[{Yuan et~al.(2024)Yuan, Song, Chen, Tan, Shen, Ren, Li, and Yang}]{yuan2024easytool}
Siyu Yuan, Kaitao Song, Jiangjie Chen, Xu~Tan, Yongliang Shen, Kan Ren, Dongsheng Li, and Deqing Yang. 2024.
\newblock {EASYTOOL}: Enhancing {LLM}-based agents with concise tool instruction.
\newblock In \emph{ICLR 2024 Workshop on Large Language Model (LLM) Agents}.

\bibitem[{Zhao et~al.(2023)Zhao, Zhou, Li, Tang, Wang, Hou, Min, Zhang, Zhang, Dong et~al.}]{zhao2023survey}
Wayne~Xin Zhao, Kun Zhou, Junyi Li, Tianyi Tang, Xiaolei Wang, Yupeng Hou, Yingqian Min, Beichen Zhang, Junjie Zhang, Zican Dong, et~al. 2023.
\newblock A survey of large language models.
\newblock \emph{arXiv preprint arXiv:2303.18223}.

\bibitem[{Zhou et~al.(2024)Zhou, Zhu, Jin, and Dou}]{10.1145/3589334.3645482}
Yujia Zhou, Qiannan Zhu, Jiajie Jin, and Zhicheng Dou. 2024.
\newblock Cognitive personalized search integrating large language models with an efficient memory mechanism.
\newblock In \emph{Proceedings of the ACM Web Conference 2024}, page 1464–1473. Association for Computing Machinery.

\bibitem[{Zhuang et~al.(2024)Zhuang, Chen, Yu, Mitra, Bursztyn, Rossi, Sarkhel, and Zhang}]{zhuang2024toolchain}
Yuchen Zhuang, Xiang Chen, Tong Yu, Saayan Mitra, Victor Bursztyn, Ryan~A. Rossi, Somdeb Sarkhel, and Chao Zhang. 2024.
\newblock Toolchain*: Efficient action space navigation in large language models with a* search.
\newblock In \emph{The Twelfth International Conference on Learning Representations}.

\end{thebibliography}

\appendix

\section{Details of Benchmark Construction}
\begin{figure}[htbp]
    \centering
    \includegraphics[width=1.0\columnwidth]{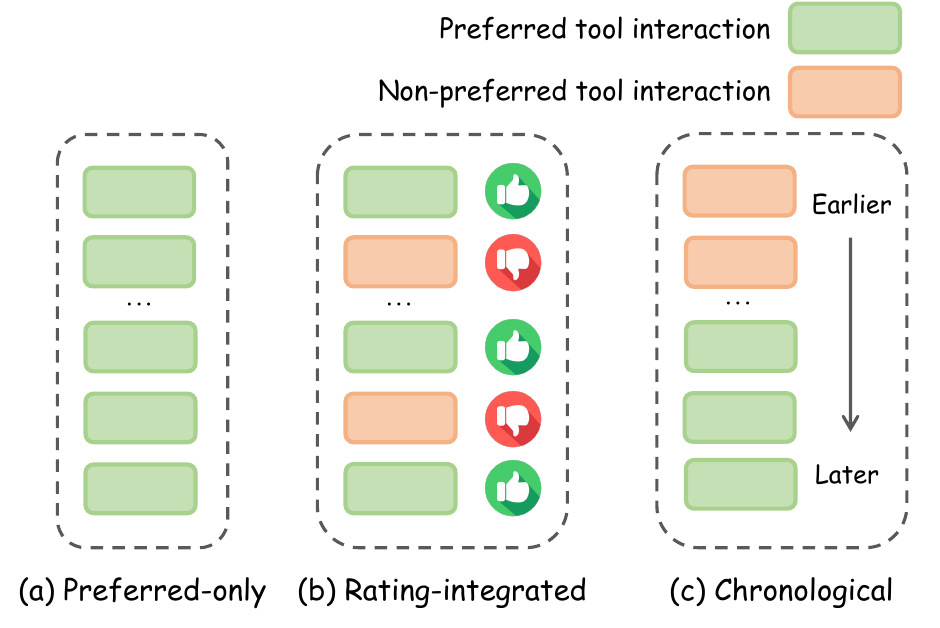}
    \caption{Illustration of three types of the user's interaction history.}
    \label{fig:history}
\vspace{-1em}
\end{figure}

We provide the illustration of three types of the user's interaction history in Figure~\ref{fig:history}.

\section{Implementation details}
To train \framework, we fine-tune the LLaMA-3.1-8B model with LoRA
and a warm-up ratio of $0.1$ in the SFT stage. 
The learning rate is set to $1e{-4}$ with a batch size of $16$ per GPU. 
In the DPO stage, the learning rate is set to $1e{-6}$ and the balancing factor $\beta$ is set to $0.1$ with a batch size of $32$.
We have trained the model several times to ensure the improvement is not randomly achieved and present the mid one. 
For evaluation, we set the number of candidate tools $N$ to $10$ and the temperature to 0.1 to reduce randomness. 
Since the maximum context length varies in different LLMs, we constrain the context window to 4000 tokens. The experiments on closed-source LLMs are fulfilled by APIs of OpenAI and those on open-source LLMs are conducted on NVIDIA A6000 GPUs with 48 GB of memory. 

\definecolor{lightgray}{RGB}{240, 240, 240}
\lstdefinestyle{prompt}{
    basicstyle=\ttfamily\fontsize{7pt}{8pt}\selectfont,
    frame=none,
    breaklines=true,
    backgroundcolor=\color{lightgray},
    breakatwhitespace=true,
    breakindent=0pt,
    escapeinside={(*@}{@*)},
    numbers=none,
    numbersep=5pt,
    xleftmargin=5pt,
}
\tcbset{
  aibox/.style={
  }
}
\newtcolorbox{AIbox}[2][]{aibox, title=#2,#1}

\section{Prompt Details}
The prompt templates in for tool-use example generation and tool attributes understanding are shown in Figure~\ref{fig:prompt_example} and Figure~\ref{fig:prompt_attributes}. The prompt templates for interaction history generation across three types are shown in Figure~\ref{fig:prompt_history(p)}, Figure~\ref{fig:prompt_history(r)}, and Figure~\ref{fig:prompt_history(c)}.
The prompt template for instruction generation is shown in Figure~\ref{fig:prompt_instruction}.

\begin{figure*}[!ht] 
\vspace{-5mm}
\begin{AIbox}{Prompt for Tool-use Example Generation}
{\bf Prompt:} \\
{
Given a tool documentation as input, your task is to output an example for using this tool, including a simulated user instruction and parameters for calling the tool. The output example should be in JSON format: \{``instruction'': xx, ``parameters'': xx\}
\clearpage
Here is a demonstration:

Input:
\begin{lstlisting}[style=prompt]
{
    "tool_name": "<Text_Analysis>.<Spellout>.<Languages>",
    "tool_desciption": "List ISO 639 languages",
    "required_parameters": [],
    "optional_parameters": [
        {
            "name": "nameFilter",
            "type": "STRING",
            "description": "Filter as \"contains\" by language name",
            "default": ""
        }
    ]
}
\end{lstlisting}
Output:
\begin{lstlisting}[style=prompt]
{
    "instruction": "I want to filter the list of languages by English",
    "parameters": {
        "nameFilter": "English"
    }
}
\end{lstlisting}
Now you will be given the tool documentation, please generate the tool-use example. 

Begin!
}
\end{AIbox} 
\caption{The prompt for tool-use example generation.}
\label{fig:prompt_example}
\vspace{-5mm}
\end{figure*}

\begin{figure*}[!ht] 
\vspace{-5mm}
\begin{AIbox}{Prompt for Tool Attributes Understanding}
{\bf Prompt:} \\
{
Given a tool documentation and the corresponding tool-use example as input, your task is to understand the tool attributes thoroughly. Then generate two descriptions about the functionality and non-functional attributes of the tool respectively. 
\clearpage
Here is a demonstration:

Input:
\begin{lstlisting}[style=prompt]
Tool documentation:
{
    "tool_name": "<Commerce>.<Face Compare>.<GET Call>",
    "tool_desciption": "Used to fetch results using the request id received in responses.",
    "required_parameters": [
        {
            "name": "request_id",
            "type": "STRING",
            "description": "",
            "default": "76d1c748-51ed-435b-bcd8-3d9c9d3eb68a"
        }
    ],
Tool-use example:
{
    "instruction": "I want to use the request id '76d1c748-51ed-435b-bcd8-3d9c9d3eb68a' to fetch the result",
    "parameters": {
        "request_id": "76d1c748-51ed-435b-bcd8-3d9c9d3eb68a"
    }
}
\end{lstlisting}
Output:
\begin{lstlisting}[style=prompt]
Functionality: Fetches API results based on the request ID received in previous responses.
Non-functional attributes: Designed for commerce applications, used in face comparison scenarios.
\end{lstlisting}
Now you will be given the tool documentation and the tool-use example, generate two short phrases to describe the two types of attributes. 

Begin!
}
\end{AIbox} 
\caption{The prompt for tool attributes understanding.}
\label{fig:prompt_attributes}
\vspace{-5mm}
\end{figure*}

\begin{figure*}[!ht] 
\vspace{-5mm}
\begin{AIbox}{Prompt for Interaction History (Preferred-only) Generation}
{\bf Prompt:} \\
{
Given a list of tools preferred by a user as input, your task is to simulate the user's interaction history based on these tools. You should output a sequence of tool-usage interactions, each consisting of a simulated user instruction and a tool call to fulfill that instruction. The interaction sequence should be a list in JSON format: 
\begin{lstlisting}[style=prompt]
[
    {
        "instruction": xx,
        "tool_call": {
            "tool_name": xx, 
            "parameters": xx
        }
    }, ...
]
\end{lstlisting}
Now you will be given the tools, please generate the interaction sequence. 

Begin!
}
\end{AIbox} 
\caption{The prompt for interaction history (preferred-only) generation.}
\label{fig:prompt_history(p)}
\vspace{-5mm}
\end{figure*}

\begin{figure*}[!ht] 
\vspace{-5mm}
\begin{AIbox}{Prompt for Interaction History (Rating-integrated) Generation}
{\bf Prompt:} \\
{
Given a list of tools preferred by a user and a list of tools not preferred as input, your task is to simulate the user's interaction history based on these two lists. You should output a sequence of tool-usage interactions, each consisting of a simulated user instruction, a tool call to fulfill that instruction, and a binary rating reflecting the user's satisfaction with the tool call. The interaction sequence should be a list in JSON format: 
\begin{lstlisting}[style=prompt]
[
    {
        "instruction": xx,
        "tool_call": {
            "tool_name": xx, 
            "parameters": xx
        },
        "rating": 1 or 0,
    }, ...
]
\end{lstlisting}
Now you will be given the two lists of tools, please generate the interaction sequence. 

Begin!
}
\end{AIbox} 
\caption{The prompt for interaction history (rating-integrated) generation.}
\label{fig:prompt_history(r)}
\vspace{-5mm}
\end{figure*}

\begin{figure*}[!ht] 
\vspace{-5mm}
\begin{AIbox}{Prompt for Interaction History (Chronological) Generation}
{\bf Prompt:} \\
{
Given a list of tools preferred by a user and a list of tools not preferred as input, your task is to simulate the user's interaction history based on these two lists. You should output a sequence of tool-usage interactions, each consisting of a simulated user instruction, a tool call to fulfill that instruction. The interactions should be organized in time order to reflect changes in user preferences over time, i.e., the more recent tool-usage interactions are more preferred by the user, while earlier interactions are less preferred. The interaction sequence should be a list in JSON format: 
\begin{lstlisting}[style=prompt]
[
    {
        "instruction": xx,
        "tool_call": {
            "tool_name": xx, 
            "parameters": xx
        }
    }, ...
]
\end{lstlisting}
Now you will be given the two lists of tools, please generate the interaction sequence. 

Begin!
}
\end{AIbox} 
\caption{The prompt for interaction history (chronological) generation.}
\label{fig:prompt_history(c)}
\vspace{-5mm}
\end{figure*}

\begin{figure*}[!ht] 
\begin{AIbox}{Prompt for Instruction Generation}
{\bf Prompt:} \\
{
Given a user's interaction history and a tool documentation as input, your task is to generate a simulated user instruction which can be fulfilled by calling the tool with parameters. The generated output should be in JSON format: 
\begin{lstlisting}[style=prompt]
{
    "instruction": xx,
    "parameters": xx
}

\end{lstlisting}
Remember, tool name is strictly prohibited from appearing in the generated instruction. 
Now you will be given the user's interaction history and tool documentation, please generate the output. 

Begin!
}
\end{AIbox} 
\caption{The prompt for instruction generation.}
\label{fig:prompt_instruction}
\end{figure*}

\label{sec:appendix}


\end{document}